\documentclass[letterpaper, 10 pt, conference]{ieeeconf}  %

\IEEEoverridecommandlockouts                              %

\usepackage{graphics} %
\usepackage{epsfig} %
\usepackage{mathptmx} %
\usepackage{times} %
\usepackage{amsmath} %
\usepackage{amssymb}  %
\usepackage{url}

\usepackage{booktabs}
\usepackage{multicol}
\usepackage{multirow}
\usepackage[bookmarks=true]{hyperref}
\newcommand{\modelname}[1]{RFPO}

\title{\LARGE \bf
\modelname{}: Rectified Flow Policy Optimization for Embodied Control
}

\author{}

\author{
    \textbf{Ting Huang}$^{1*}$\quad
    \textbf{Lisiyu Pan}$^{1*}$\quad 
    \textbf{Haoyu Wang}$^{1*}$\quad 
    \textbf{Zeyu Zhang}$^{1*\dag}$\quad
    \textbf{Siyuan Qian}$^{1}$\\
    \textbf{Yanjun Li}$^{1}$\quad
    \textbf{Yandong Guo}$^{2}$\quad
    \textbf{Boxin Shi}$^{1}$\quad
    \textbf{Hao Tang}$^{1\ddag}$ \vspace{0.1cm}\\
    $^1$School of Computer Science, Peking University\quad
    $^2$AI$^2$ Robotics\vspace{0.05cm}\\
    \small $^*$Equal contribution. $^\dag$Project lead.
    $^\ddag$Corresponding author: bjdxtanghao@gmail.com
}

\begin{document}

\maketitle
\thispagestyle{empty}
\pagestyle{empty}

\begin{abstract}
Flow-based policies provide an expressive framework for continuous robot control, but their iterative ODE integration incurs substantial inference cost.
Naively reducing the integration budget can severely degrade control, since policies optimized under full-step execution are not explicitly constrained to remain reliable under coarse numerical integration.
We refer to this mismatch as the \emph{few-step discretization gap}.
To address this problem, we introduce \modelname{}, a flow-policy optimization framework for reliable few-step execution.
Reward-aware online Reflow rectifies student-induced transport paths during on-policy learning, making the resulting policy more robust to coarse integration.
A frozen Gaussian PPO controller supplies complementary action-space supervision at full and intermediate integration budgets, while the deployed policy remains a single flow student executed with one Euler step.
Across Unitree Go2, Boston Dynamics Spot, Unitree H1, and Unitree G1, \modelname{} consistently preserves full-step control performance under one-step execution, with one-step returns remaining within $2.4\%$ of their corresponding 64-step values across both zero and random initialization.
On Unitree Go2, one-step execution retains \textbf{98.5\%} of the 64-step reward while reducing onboard mean inference latency from $4.39$~ms to $0.08$~ms, yielding a \textbf{54.9$\times$ speedup}.
Real-robot experiments further validate stable one-step locomotion.
Code:~\url{https://github.com/AIGeeksGroup/RFPO}.
Website:~\url{https://aigeeksgroup.github.io/RFPO}
\end{abstract}

\section{Introduction}
Learning expressive and efficient continuous-control policies is a central problem in robotics.
Conventional reinforcement learning typically parameterizes actions with simple distributions such as diagonal Gaussians, which are easy to optimize but can be restrictive when multiple control strategies are valid under the same observation.
Generative policies provide a more expressive alternative and have shown strong performance in visuomotor and geometry-aware robot control~\cite{chi2025diffusion,ze20243d}.
In particular, flow matching~\cite{lipman2022flow} and rectified flow~\cite{liu2022flow} model action generation as an ODE that transports simple noise distributions into continuous actions.
Flow-based policies have since been adopted for efficient robot manipulation~\cite{zhang2025flowpolicy} and large-scale vision-language-action models such as $\pi_0$~\cite{black2024pi_0}, highlighting their potential as an expressive policy class for robot control.

Beyond demonstration-supervised learning, recent work has increasingly optimized generative policies directly from environment rewards.
Diffusion-based approaches such as DPPO~\cite{ren2025diffusion}, DIME~\cite{celik2025dime}, and GenPO~\cite{ding2026genpo} extend reinforcement learning to iterative generative action distributions.
For flow-based policies, ReinFlow~\cite{zhang2026reinflow} enables online fine-tuning through stochastic flow dynamics, while Flow Q-Learning~\cite{park2025flow} and SAC Flow~\cite{zhang2026sac} explore offline and off-policy optimization.
FPO~\cite{mcallister2026flow} constructs an on-policy surrogate directly from conditional flow-matching losses, and FPO++~\cite{yi2026flow} further improves this formulation with per-sample flow ratios and asymmetric trust-region updates.
These advances make reward-driven optimization of expressive flow policies increasingly practical.

\begin{figure}[t]
    \centering
    \includegraphics[width=\linewidth]{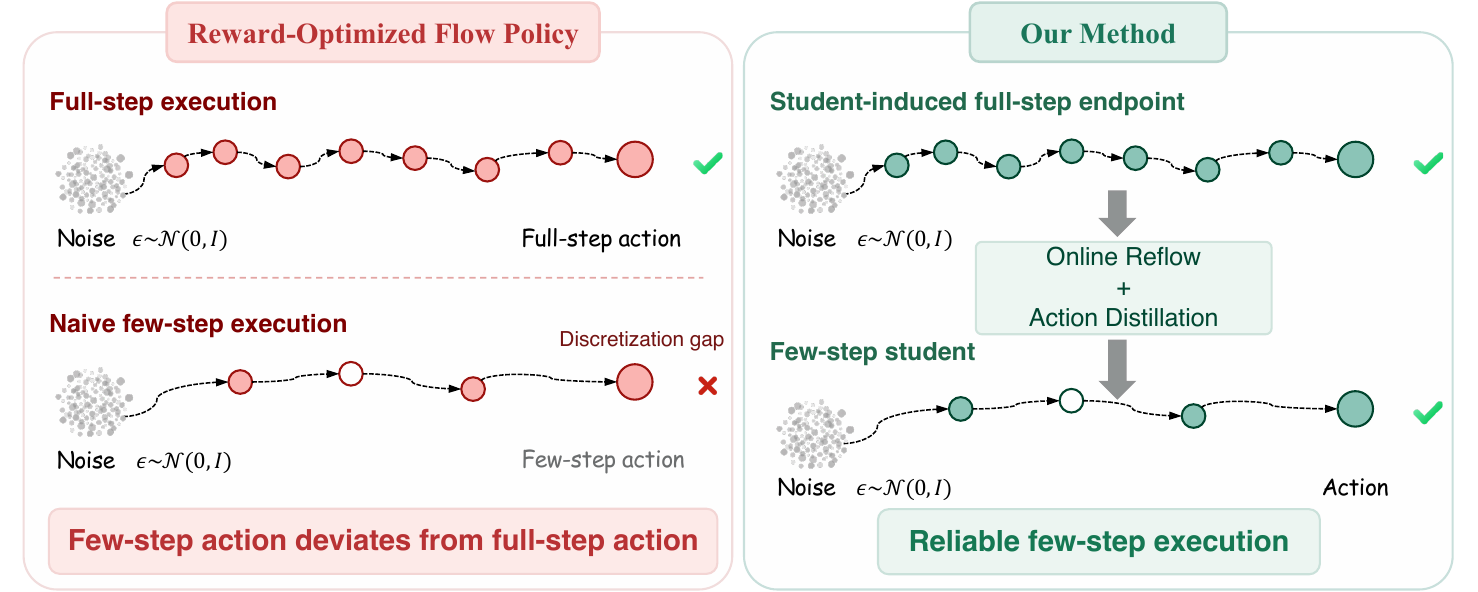}
    \caption{\textbf{Few-step execution of reward-optimized flow policies.}
    Naive step reduction can create a discretization gap between full-step and few-step actions.
    \modelname{} rectifies student-induced transport paths and uses complementary action-space supervision to enable reliable few-step execution.}
    \label{fig:motivation}
    \vspace{-0.7cm}
\end{figure}
However, high-reward flow policies are not necessarily reliable when the numerical integration budget is aggressively reduced.
Existing flow-policy RL objectives optimize control return under the training integration budget but do not explicitly constrain the resulting vector field to remain numerically reliable under coarse integration.
Consequently, reducing the number of Euler steps can substantially change the generated action and lead to severe control degradation, as illustrated in Fig.~\ref{fig:motivation}.
We refer to this mismatch between full-step optimization and low-budget execution as the \emph{few-step discretization gap}.
Although Reflow~\cite{liu2022flow}, consistency learning~\cite{song2023consistency}, fast ODE solvers, and policy distillation~\cite{prasad2024consistency,zhang2025flowpolicy} have enabled efficient few-step generation, these techniques are primarily developed for generative modeling, imitation learning, or post-hoc policy acceleration rather than jointly with on-policy reward optimization.
This motivates directly shaping reward-optimized flow policies for reliable few-step execution.

To address this problem, we introduce \modelname{}, a flow-policy optimization framework for reliable few-step control.
The key mechanism is reward-aware online Reflow, which repeatedly constructs student-induced full-step endpoints and rectifies their transport paths during on-policy learning, making the learned vector field more amenable to coarse numerical integration.
A frozen Gaussian PPO controller supplies complementary action-space supervision at full and intermediate integration budgets, anchoring the quality of the learned actions without directly supervising one-step execution.
The original on-policy objective is retained throughout training, and all training-only components are removed at deployment, leaving a single flow student executed with one Euler step.
We evaluate \modelname{} across two quadruped and two humanoid embodiments: Unitree Go2, Boston Dynamics Spot, Unitree H1, and Unitree G1.
Across all four embodiments and both zero and random initialization, one-step execution remains within $2.4\%$ of the corresponding 64-step return, whereas FPO++ and distillation-only baselines exhibit substantial degradation under aggressive step reduction.
On Unitree Go2, \modelname{} retains \textbf{98.5\%} of its 64-step reward while reducing onboard mean inference latency from $4.39$~ms to $0.08$~ms, yielding a \textbf{54.9$\times$ speedup}.
Real-robot experiments further demonstrate stable one-step locomotion across forward, lateral, and yaw commands.

Our contributions are threefold:
\begin{itemize}
    \item We identify the \emph{few-step discretization gap} in reward-optimized flow policies, where aggressive reduction of the numerical integration budget can severely degrade control despite strong full-step performance.

    \item We introduce \modelname{}, which uses reward-aware student-induced online Reflow to regularize transport geometry during on-policy learning, together with complementary multi-budget action-space supervision from a frozen Gaussian PPO controller.

    \item Extensive evaluations across two quadruped and two humanoid embodiments demonstrate robust one-step execution.
    On Unitree Go2, \modelname{} retains \textbf{98.5\%} of 64-step reward with a \textbf{54.9$\times$ onboard inference speedup} and transfers to real-robot locomotion.
\end{itemize}

\begin{figure*}[t]
    \centering
    \small
    \includegraphics[width=\linewidth]{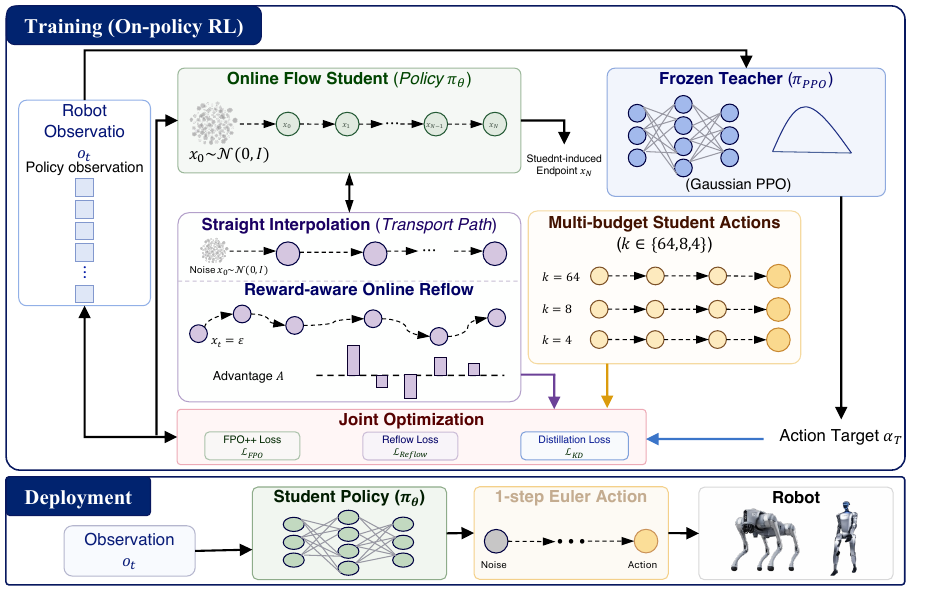}
    \caption{\textbf{Overview of \modelname{}.}
    During on-policy training, student-induced full-step endpoints are rectified by reward-aware online Reflow, while multi-budget student actions are distilled toward a frozen PPO controller.
    An adaptive budget objective further regularizes the fidelity--compute trade-off.
    At deployment, all auxiliary components are removed and only the flow student is retained for one-step Euler execution.}
    \label{fig:method}
    \vspace{-0.6cm}
\end{figure*}

\section{Related Work}
\noindent\textbf{Generative and flow policies for robot control.}
Generative policies provide an expressive alternative to Gaussian actors for modeling multimodal robot behaviors.
Diffusion Policy~\cite{chi2025diffusion} and 3D Diffusion Policy~\cite{ze20243d} demonstrate strong visuomotor and geometry-aware manipulation performance using iterative generative action models.
Flow matching~\cite{lipman2022flow} and rectified flow~\cite{liu2022flow} provide an ODE-based alternative and have been adopted by robot policies such as RF-Policy~\cite{hu2023rf}, FlowPolicy~\cite{zhang2025flowpolicy}, and ManiFlow~\cite{yan2025maniflow}, as well as large-scale vision-language-action models such as $\pi_0$~\cite{black2024pi_0}.
Recent work has further accelerated iterative robot policies through consistency learning and policy distillation~\cite{prasad2024consistency,wang2025onestep}.
These methods are predominantly trained from demonstrations or supervised robot data, whereas our work studies few-step execution of flow policies optimized directly from environment rewards.

\noindent\textbf{Reinforcement learning with generative policies.}
Generative models have increasingly been incorporated into planning and reinforcement learning.
Diffusion-based methods have been used for trajectory planning and expressive offline policies~\cite{janner2022diffuser,wang2023diffusion,kang2023efficient}, while Q-score matching connects diffusion-policy scores with reward-derived action gradients~\cite{psenka2023learning}.
DPPO~\cite{ren2025diffusion}, DIME~\cite{celik2025dime}, and GenPO~\cite{ding2026genpo} further optimize iterative generative policies with reinforcement learning.
For flow-based policies, ReinFlow~\cite{zhang2026reinflow} studies online fine-tuning, while Flow Q-Learning~\cite{park2025flow} and SAC Flow~\cite{zhang2026sac} consider offline or off-policy optimization.
FPO~\cite{mcallister2026flow} and FPO++~\cite{yi2026flow} instead develop on-policy objectives directly from conditional flow-matching losses.
These methods primarily optimize control return, while our work addresses the complementary problem of preserving policy behavior when the numerical integration budget is aggressively reduced.

\noindent\textbf{Few-step generative policy acceleration.}
Reducing the inference cost of iterative generative models has motivated both solver-based and learning-based acceleration.
Training-free approaches such as DDIM~\cite{song2021denoising} and DPM-Solver~\cite{lu2022dpm} reduce the number of numerical integration steps, while learning-based approaches include progressive distillation~\cite{salimans2022progressive}, consistency models~\cite{song2023consistency}, consistency trajectory models~\cite{kim2024consistency}, improved consistency training~\cite{song2024improved}, and Shortcut Models~\cite{frans2025one}.
For flow-based models, Reflow~\cite{liu2022flow} and InstaFlow~\cite{liu2024instaflow} reshape transport trajectories to support coarse or one-step generation.
Related ideas have been transferred to robot policies through Consistency Policy~\cite{prasad2024consistency}, One-Step Diffusion Policy~\cite{wang2025onestep}, FlowPolicy~\cite{zhang2025flowpolicy}, and ManiFlow~\cite{yan2025maniflow}.
These approaches largely target generative modeling, imitation learning, or post-hoc policy acceleration.
In contrast, \modelname{} performs student-induced Reflow during on-policy reward optimization, directly regularizing the transport geometry encountered by the evolving policy.
A frozen deployable controller provides complementary action-space supervision at full and intermediate integration budgets rather than direct one-step supervision.

\noindent\textbf{Learning-based legged locomotion and sim-to-real.}
Deep reinforcement learning has enabled agile legged locomotion through simulation-based training, robust policy learning, and massively parallel simulation~\cite{tan2018sim,hwangbo2019agile,rudin2021learning}.
Subsequent work has improved real-world robustness through rapid adaptation and perceptive locomotion~\cite{kumar2021rma,miki2022perceptive}.
These studies primarily address dynamics mismatch, terrain variation, and online adaptation, whereas our work focuses on a complementary deployment bottleneck: preserving the behavior of reward-optimized flow policies under aggressive reductions of the numerical integration budget.

\section{Method}

\subsection{Overview and Problem Formulation}
We consider a conditional flow policy $v_\theta(o,t,x)$ that maps a robot observation $o$ to a continuous action $a\in\mathbb{R}^{d_a}$.
We follow the convention of the underlying flow-policy optimizer, where $t=1$ denotes the noise endpoint and $t=0$ denotes the action endpoint.
Given Gaussian noise $\epsilon\sim\mathcal{N}(0,I)$ and an action $a$, conditional flow matching defines the linear probability path
\begin{equation}
x_t=t\epsilon+(1-t)a,
\qquad
u^\star=\epsilon-a .
\label{eq:cfm_path}
\end{equation}

At inference time, the learned velocity field is solved numerically from $t=1$ to $t=0$.
We denote the action obtained with $k$ Euler steps by
\begin{equation}
a^{(k)}
=
\mathcal{T}_{v_\theta}^{(k)}(o,\epsilon),
\label{eq:k_step_action}
\end{equation}
where the standard full budget is $N=64$.

We build on FPO++~\cite{yi2026flow} and denote its original on-policy actor--critic objective by $\mathcal{L}_{\mathrm{FPO}}$.
This objective optimizes control performance at the training integration budget, but does not explicitly shape the vector field for coarse numerical execution.
As a result, the action produced with a small budget $k\ll N$ can deviate substantially from its full-budget counterpart even when both are generated by the same policy.
We refer to this mismatch as the \emph{few-step discretization gap}.

As illustrated in Fig.~\ref{fig:method}, \modelname{} improves the numerical deployability of the learned flow through two complementary training signals.
Reward-aware online Reflow shapes student-induced transport paths toward straighter couplings, while multi-budget action distillation anchors the resulting actions to a reliable Gaussian controller.
Auxiliary zero-start sampling exposes the student to deterministic deployment, and an adaptive compute regularizer encourages action consistency across a range of solver budgets.
All auxiliary components are used only during training; deployment retains a single flow student executed with one Euler step.

\subsection{Reward-Aware Online Reflow}
The geometry of a learned flow directly affects the error introduced by coarse numerical solvers.
We therefore apply Reflow~\cite{liu2022flow} to trajectories induced by the current student during on-policy training.

For each minibatch observation $o_i$, we draw $n=4$ auxiliary initializations.
To cover both stochastic training and deterministic deployment, each initialization is sampled as
\begin{equation}
\epsilon_i^j=
\begin{cases}
0, & \text{with probability } p_0,\\
\xi_i^j,\quad \xi_i^j\sim\mathcal{N}(0,I),
& \text{otherwise},
\end{cases}
p_0=0.25 .
\label{eq:aux_init}
\end{equation}
This mixture is restricted to the auxiliary Reflow and distillation losses; on-policy rollout collection retains Gaussian initialization.

The current student first produces a full-budget endpoint
\begin{equation}
\bar{a}_i^j
=
\operatorname{sg}
\left[
\mathcal{T}_{v_\theta}^{(N)}
(o_i,\epsilon_i^j)
\right],
\label{eq:reflow_endpoint}
\end{equation}
where $\operatorname{sg}[\cdot]$ denotes stop-gradient.
Hence, the coupling used by Reflow is induced by the evolving student itself, rather than by the external PPO teacher.

For each pair $(\epsilon_i^j,\bar{a}_i^j)$, we sample a flow time $t_i^j$ with the bounded CFM sampler used in our implementation, which is approximately uniform over $[0.005,0.995]$.
The corresponding straight path and target velocity are
\begin{equation}
\tilde{x}_{t}^{i,j}
=
t_i^j\epsilon_i^j
+
(1-t_i^j)\bar{a}_i^j,
\qquad
\tilde{u}^{i,j}
=
\epsilon_i^j-\bar{a}_i^j .
\label{eq:reflow_path}
\end{equation}

Following the CFM implementation, we use the action-dimensional square-root reduction
\begin{equation}
\rho(e)
=
\frac{\|e\|_2^2}{\sqrt{d_a}},
\label{eq:sqrt_reduction}
\end{equation}
yielding the Reflow regression loss
\begin{equation}
\ell_{i,j}^{\mathrm{RF}}
=
\rho\!\left(
v_\theta(o_i,t_i^j,\tilde{x}_{t}^{i,j})
-
\tilde{u}^{i,j}
\right).
\label{eq:reflow_sample}
\end{equation}

We use the on-policy advantage to focus rectification on behavior that is useful for control.
For the advantage estimate $A_i$, we define
\begin{equation}
w_i=[A_i-\tau]_+,
\qquad \tau=0.
\label{eq:reflow_weight}
\end{equation}
When positive-advantage samples are present, their normalized weights are
\begin{equation}
\tilde{w}_i
=
\frac{w_i}{\sum_m w_m}.
\label{eq:reflow_weight_norm}
\end{equation}
If all advantages in the minibatch are non-positive, uniform weights are used.

The resulting reward-aware Reflow objective is
\begin{equation}
\mathcal{L}_{\mathrm{reflow}}
=
\frac{1}{n}
\sum_i
\tilde{w}_i
\sum_{j=1}^{n}
\ell_{i,j}^{\mathrm{RF}},
\qquad n=4.
\label{eq:reflow_loss}
\end{equation}
This objective continuously aligns the evolving student with straighter transport couplings while retaining the reward-driven update of the original on-policy objective.

\subsection{Multi-Budget Action Distillation}
Transport rectification improves numerical behavior, while control quality also depends on the action endpoint represented by the learned field.
We therefore introduce a frozen Gaussian PPO controller as an independent action-space reference.
For the same observation $o$, the teacher provides
\begin{equation}
a_T
=
\operatorname{sg}
\left[
\pi_{\mathrm{PPO}}(o)
\right].
\label{eq:teacher_action}
\end{equation}
The PPO controller is used only to define action targets and does not participate in the construction of the Reflow couplings in Eq.~\eqref{eq:reflow_endpoint}.

The same flow student is evaluated at the full and intermediate budgets
\begin{equation}
\mathcal{K}_{\mathrm{KD}}
=
\{64,8,4\},
\label{eq:kd_budgets}
\end{equation}
and its actions are trained toward the PPO reference through
\begin{equation}
\mathcal{L}_{\mathrm{KD}}
=
\frac{1}{|\mathcal{K}_{\mathrm{KD}}|}
\sum_{k\in\mathcal{K}_{\mathrm{KD}}}
\left\|
\mathcal{T}_{v_\theta}^{(k)}(o,\epsilon)
-
a_T
\right\|_2^2 .
\label{eq:kd_loss}
\end{equation}

The two objectives therefore constrain different aspects of the policy.
Online Reflow shapes the transport geometry encountered by coarse numerical solvers, whereas action distillation preserves a reliable action-space reference across full and intermediate budgets.
Notably, $k=1$ is not included in $\mathcal{K}_{\mathrm{KD}}$; one-step behavior is consequently obtained without direct one-step teacher supervision.

\subsection{Auxiliary Adaptive Compute Regularization}
We further expose the student to heterogeneous numerical budgets during training.
A lightweight predictor $\phi_\psi(o)$ defines a categorical distribution over
\begin{equation}
\mathcal{K}_{\mathrm{ada}}
=
\{1,4,8,16,64\},
\label{eq:adaptive_budgets}
\end{equation}
with
\begin{equation}
p_\psi(k|o)
=
\operatorname{softmax}
\left(
\phi_\psi(o)
\right)_k .
\label{eq:adaptive_distribution}
\end{equation}

The fidelity branch uses the highest-probability budget
\begin{equation}
\hat{k}(o)
=
\arg\max_{k\in\mathcal{K}_{\mathrm{ada}}}
p_\psi(k|o).
\label{eq:adaptive_selection}
\end{equation}
We use the detached full-budget action under zero initialization,
\begin{equation}
a_0^{(N)}
=
\operatorname{sg}
\left[
\mathcal{T}_{v_\theta}^{(N)}(o,0)
\right],
\label{eq:adaptive_reference}
\end{equation}
as the reference and define
\begin{equation}
\mathcal{L}_{\mathrm{fid}}
=
\mathbb{E}_o
\left[
\left\|
\mathcal{T}_{v_\theta}^{(\hat{k}(o))}(o,0)
-
a_0^{(N)}
\right\|_2^2
\right].
\label{eq:adaptive_fidelity}
\end{equation}

A complementary compute term penalizes the expected normalized solver budget,
\begin{equation}
\mathcal{L}_{\mathrm{cost}}
=
\mathbb{E}_o
\left[
\sum_{k\in\mathcal{K}_{\mathrm{ada}}}
p_\psi(k|o)
\frac{k}{K_{\max}}
\right],
\qquad K_{\max}=64 .
\label{eq:adaptive_cost}
\end{equation}

Following the training implementation, the auxiliary regularizer is
\begin{equation}
\mathcal{L}_{\mathrm{ada}}
=
\mathcal{L}_{\mathrm{fid}}
+
\alpha\mathcal{L}_{\mathrm{cost}}
-
\beta
\mathbb{E}_o
\left[
[A(o)]_+
\right],
\label{eq:adaptive_loss}
\end{equation}
where $\alpha=1$ and $\beta=0.1$.
This regularizer broadens the solver budgets encountered during training and encourages low-cost execution while maintaining consistency with the full-budget student.
The budget predictor is used only during optimization and is not required at deployment.

\subsection{Joint Optimization and Deployment}
The final training objective combines the original on-policy objective with the auxiliary flow and action regularizers:
\begin{equation}
\mathcal{L}
=
\mathcal{L}_{\mathrm{FPO}}
+
\lambda_r\mathcal{L}_{\mathrm{reflow}}
+
\lambda_{\mathrm{KD}}\mathcal{L}_{\mathrm{KD}}
+
\lambda_a\mathcal{L}_{\mathrm{ada}},
\label{eq:total_loss}
\end{equation}
where
\begin{equation}
\lambda_r=1,
\qquad
\lambda_{\mathrm{KD}}=0.1,
\qquad
\lambda_a=0.1 .
\end{equation}

Training follows the on-policy rollout and advantage-estimation procedure of FPO++.
Student-generated Reflow endpoints and PPO action targets are detached from their respective target branches.
Gaussian initialization is retained for policy rollouts, while the auxiliary losses additionally receive zero-start samples through Eq.~\eqref{eq:aux_init}.

Deployment requires neither the PPO teacher nor the budget predictor.
Given observation $o_t$, we set $\epsilon=0$ and execute the learned flow student with a single Euler step,
\begin{equation}
a_t
=
\mathcal{T}_{v_\theta}^{(1)}(o_t,0).
\label{eq:deployment}
\end{equation}
Thus, the deployed controller consists only of the learned student policy, with no auxiliary teacher or adaptive-compute module.

\begin{table*}[t]
    \centering
    \small
    \caption{\textbf{Cross-embodiment few-step control.}
    Average episodic reward under 64-step and one-step Euler execution across two quadruped and two humanoid embodiments.
    PPO uses a single deterministic forward pass and therefore has no integration budget.
    Reward scales differ across tasks; comparisons should be made within each embodiment. Higher is better.}
    \vspace{-0.3cm}
    \label{tab:cross_embodiment}
    \resizebox{0.6\linewidth}{!}{
    \begin{tabular}{llrrrr}
        \toprule
        \multirow{2}{*}{Embodiment} & \multirow{2}{*}{Method} & \multicolumn{2}{c}{Zero Init.} & \multicolumn{2}{c}{Random Init.} \\
        \cmidrule(lr){3-4} \cmidrule(lr){5-6}
        &  & 64-step & 1-step & 64-step & 1-step \\
        \midrule
        \multirow{4}{*}{Go2}
        & PPO~\cite{schulman2017proximal}     & \textbf{35.58} & -- & -- & -- \\
        & FPO++~\cite{yi2026flow}   & 21.29 & -187.70 & 17.66 & -220.10 \\
        & KD-only & 29.95 & -4919.70 & 27.65 & -5121.70 \\
        & \textbf{\modelname{} (Ours)} & 34.78 & \textbf{34.25} & \textbf{31.56} & \textbf{31.50} \\
        \midrule \midrule
        \multirow{4}{*}{Spot}
        & PPO~\cite{schulman2017proximal}     & \textbf{30.84} & -- & -- & -- \\
        & FPO++~\cite{yi2026flow}   & 9.29 & -404.05 & 9.31 & -195.82 \\
        & KD-only & -0.32 & $-1.5{\times}10^{5}$ & -0.77 & $-1.6{\times}10^{5}$ \\
        & \textbf{\modelname{} (Ours)} & 28.41 & \textbf{28.52}  & \textbf{24.41} & \textbf{24.98} \\
        \midrule \midrule
        \multirow{4}{*}{H1}
        & PPO~\cite{schulman2017proximal}     & 414.86 & -- & -- & -- \\
        & FPO++~\cite{yi2026flow}   & 376.67 & 18.88 & 325.16 & 22.70 \\
        & KD-only & 367.72 & 2.48 & 335.77 & 2.73 \\
        & \textbf{\modelname{} (Ours)} & \textbf{417.50} & \textbf{413.78} & \textbf{404.42} & \textbf{406.36} \\
        \midrule \midrule
        \multirow{4}{*}{G1}
        & PPO~\cite{schulman2017proximal}     & \textbf{56.37} & -- & -- & -- \\
        & FPO++~\cite{yi2026flow}   & 18.28 & -1.51 & 4.00 & -2.82 \\
        & KD-only & 32.89 & -4.39 & 19.97 & -4.68 \\
        & \textbf{\modelname{} (Ours)} & 54.30 & \textbf{55.24} & \textbf{52.35} & \textbf{52.03} \\
        \bottomrule
    \end{tabular}
    }
\end{table*}

\section{Experiments}

\subsection{Experimental Setup}

\noindent\textbf{Environment and task.}
We evaluate \modelname{} on velocity-control tasks spanning two quadruped and two humanoid embodiments: Unitree Go2, Boston Dynamics Spot, Unitree H1, and Unitree G1 (29 DoF).
Each policy is trained and evaluated using the corresponding embodiment-specific observation space, reward function, and low-level control interface.
For each embodiment, the PPO teacher and all flow-policy variants use the same task definition and domain-randomization protocol.
For the real-robot study, we use the official Unitree-Go2-Velocity task.
The Go2 policy receives a 45-D proprioceptive observation and outputs continuous joint-control targets through the official low-level PD interface.
The observation does not include base linear velocity.

\noindent\textbf{Compared methods.}
We compare \modelname{} with the Gaussian PPO controller~\cite{schulman2017proximal} and FPO++~\cite{yi2026flow}.
We also evaluate \emph{KD-only}, which uses PPO action supervision without Reflow.
For the detailed Go2 ablation, we additionally consider uniform \emph{Reflow}, advantage-weighted \emph{Reward-aware Reflow}, and \emph{Reflow+KD}.
The complete \modelname{} uses reward-aware online Reflow, multi-budget action distillation, auxiliary zero-start training, and adaptive compute regularization.

\noindent\textbf{Integration and evaluation protocol.}
The standard full integration budget is $N=64$.
For the cross-embodiment comparison, we evaluate both 64-step and one-step Euler execution.
For the detailed Go2 analysis, we additionally sweep $N\in\{64,32,16,8,4,1\}$ without retraining for individual solver budgets.
We evaluate flow policies under both \emph{zero} initialization ($\epsilon=0$), which matches deterministic deployment, and \emph{random} initialization ($\epsilon\sim\mathcal{N}(0,I)$).
PPO uses a single deterministic policy forward pass and therefore has no integration budget.
Unless otherwise specified, evaluation uses 256 parallel environments with one episode collected per environment, yielding 256 evaluation episodes per setting.
All reported flow-policy evaluations use the EMA checkpoint (decay $0.95$ with a 500-step warm-up).

\noindent\textbf{Metrics and real-robot evaluation.}
We report average episodic reward for all embodiments.
The detailed Go2 study additionally reports the mean absolute forward-velocity tracking error $\lvert v_x-v_x^{\mathrm{cmd}}\rvert$ and fall fraction, where a fall is detected when the base height drops below $0.2$~m.
Onboard inference efficiency is measured using single-thread ONNX Runtime on the Go2 onboard CPU after 30 warm-up iterations and 100 timed evaluations; we report mean and p99 latency.
For hardware validation, command trajectories recorded on the real robot are replayed under matched inputs in simulation, and we report joint-position correlation and mean absolute error (MAE).

\begin{figure*}[t]
    \centering
    \includegraphics[width=\textwidth]{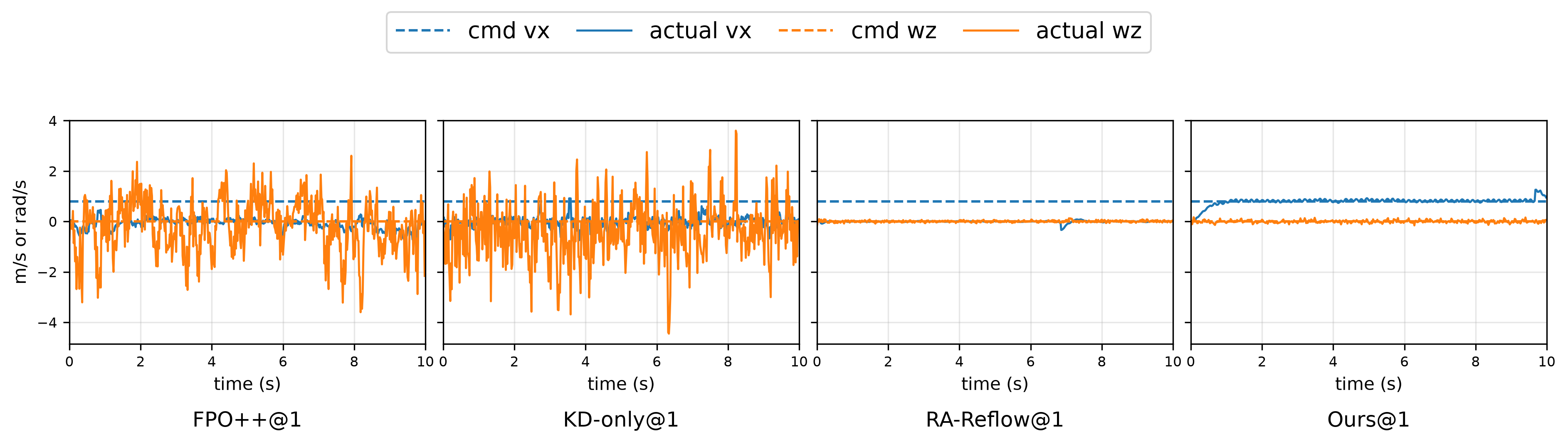}
    \caption{\textbf{Representative one-step velocity tracking on Unitree Go2.}
    Responses of FPO++, KD-only, RA-Reflow, and \modelname{} to the same forward command $v_x^{\mathrm{cmd}}=0.8$~m/s.
    FPO++ and KD-only fail to track the target under one-step execution, whereas RA-Reflow remains dynamically stable but does not follow the commanded velocity.
    \modelname{} maintains stable command tracking.}
    \label{fig:tracking_ablation}
    \vspace{0cm}
\end{figure*}

\begin{figure*}
    \centering
    \small
    \includegraphics[width=\linewidth]{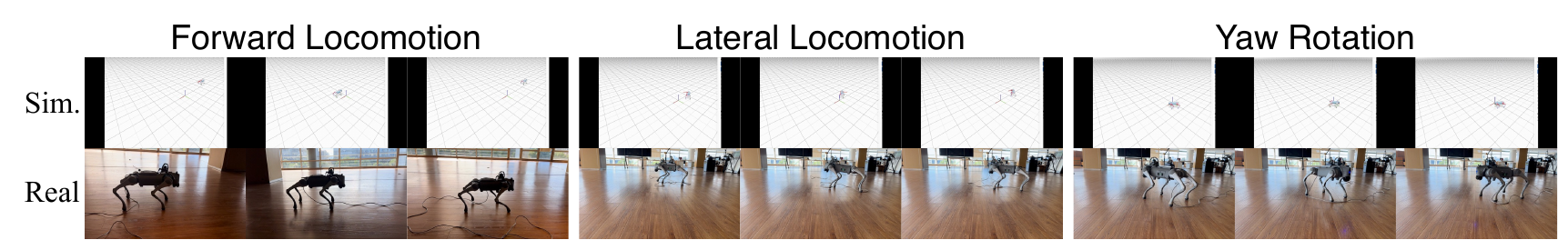}
    \caption{\textbf{Qualitative sim-to-real deployment on Unitree Go2.}
    Representative simulation and real-world executions of forward locomotion, lateral locomotion, and yaw rotation using the fixed one-step \modelname{} policy.
    Frames are sampled at comparable normalized phases of each motion sequence.}
    \label{fig:sim2real}
    \vspace{-0.6cm}
\end{figure*}

\subsection{Cross-Embodiment Few-Step Control}
Tab.~\ref{tab:cross_embodiment} evaluates the robustness of reward-optimized flow policies to aggressive integration reduction across four robot embodiments.
FPO++ exhibits a substantial few-step discretization gap on every task.
Action distillation alone does not resolve this behavior: although KD-only can improve full-step control on some embodiments, its one-step performance still degrades severely.

In contrast, \modelname{} remains remarkably insensitive to the integration budget.
Under zero initialization, reducing the solver from 64 Euler steps to one changes the average return by only $-1.5\%$, $+0.4\%$, $-0.9\%$, and $+1.7\%$ on Go2, Spot, H1, and G1, respectively.
The same trend holds under random initialization, where the corresponding 64-to-1-step changes remain within $2.4\%$ across all four embodiments.
These results show that the learned one-step behavior is neither specific to a single robot morphology nor dependent on deterministic zero initialization.

\subsection{Ablation Study}

\noindent\textbf{Effect of online Reflow.}
We first isolate the mechanisms responsible for few-step robustness on Unitree Go2.
Tab.~\ref{tab:ablation_reward} reports the zero-initialized return over the complete integration-budget sweep.
FPO++ exhibits severe degradation as the solver budget is reduced, dropping from $21.29$ at 64 steps to $-187.70$ at one step.
KD-only improves full-budget performance but remains numerically fragile, eventually collapsing to $-4919.70$ at one step.
In contrast, uniform Reflow maintains a nearly flat return profile from 64 to one step ($33.57\rightarrow32.84$), and reward-aware Reflow shows similar numerical robustness.
Combining Reflow with action-space supervision further improves the full-budget controller while preserving few-step performance.
These results identify transport rectification as the primary mechanism for robust coarse integration, with action distillation providing complementary control supervision.

\begin{table}[bhtp]
    \centering
    \small
    \caption{\textbf{Few-step robustness on Unitree Go2.}
    Average episodic reward under zero initialization across Euler integration budgets. Higher is better.}
    \label{tab:ablation_reward}
    \vspace{-0.3cm}
    \resizebox{\linewidth}{!}{
    \begin{tabular}{lrrrrrr}
        \toprule
        \multirow{2}{*}{Method} & \multicolumn{6}{c}{Euler Steps} \\
        \cmidrule(lr){2-7}
         & 64 & 32 & 16 & 8 & 4 & 1 \\
        \midrule
        FPO++      & 21.29 & 19.06 & 18.78 & 18.56 & 15.43 & -187.70 \\
        KD-only    & 29.95 & 29.59 & 29.91 & 26.63 & -398.10 & -4919.70 \\
        \midrule
        Reflow     & 33.57 & 33.50 & 33.50 & 33.45 & 33.18 & 32.84 \\
        RA-Reflow  & 32.78 & 32.98 & 33.12 & 32.59 & 33.00 & 33.40 \\
        \midrule
        Reflow+KD   & 34.65 & 34.44 & 34.77 & 34.77 & 34.92 & 34.39 \\
        \modelname{}& 34.78 & 34.43 & 34.32 & 34.67 & 34.21 & 34.25 \\
        \bottomrule
    \end{tabular}
    }
    \vspace{-0.4cm}
\end{table}

\noindent\textbf{Command tracking and stability.}
Reward preservation alone does not necessarily imply accurate low-level control.
As shown in Tab.~\ref{tab:ablation_tracking}, FPO++ and KD-only exhibit large one-step tracking errors of $0.883$ and $0.810$, accompanied by fall fractions of $0.680$ and $0.334$, respectively.
Uniform Reflow reduces the tracking error to $0.070$ with no observed falls, further confirming the importance of transport rectification for stable coarse integration.
Interestingly, RA-Reflow retains high one-step episodic reward (Tab.~\ref{tab:ablation_reward}) but incurs a tracking error of $0.805$, showing that episodic reward robustness alone does not guarantee accurate command following.
Reflow+KD and \modelname{} achieve low one-step tracking errors of $0.068$ and $0.075$, respectively, with no observed falls.
Together, these results show that Reflow provides numerical robustness, while action-space supervision complements it by preserving control fidelity.

\begin{figure*}[t]
    \centering
    \small
    \includegraphics[width=\textwidth]{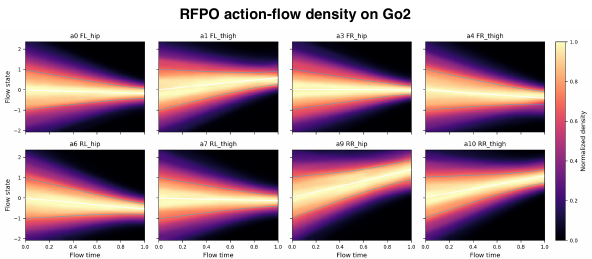}
    \caption{\textbf{Action-flow geometry of \modelname{} on Unitree Go2.}
    Normalized action-state densities over flow time for representative hip and thigh joints across all four legs.
    The learned policy exhibits concentrated and smoothly evolving transport patterns across joint dimensions.}
    \label{fig:flow_density}
    \vspace{-0.6cm}
\end{figure*}

\begin{table}[bhtp]
    \centering
    \caption{\textbf{Command tracking and stability on Unitree Go2.}
    Mean absolute forward-velocity tracking error $\lvert v_x-v_x^{\mathrm{cmd}}\rvert$ and one-step fall fraction.
    Lower is better. All evaluated 64-step flow policies have zero fall fraction.}
    \label{tab:ablation_tracking}
    \vspace{-0.3cm}
    \resizebox{0.8\linewidth}{!}{
    \begin{tabular}{lccc}
        \toprule
        \multirow{2}{*}{Method} & \multicolumn{2}{c}{Tracking Error $\downarrow$} & \multirow{2}{*}{Fall Frac. @1 $\downarrow$} \\
        \cmidrule(lr){2-3}
        & 64-step & 1-step & \\
        \midrule
        PPO     & 0.055 & -- & -- \\
        FPO++   & 0.792 & 0.883 & 0.680 \\
        KD-only & 0.178 & 0.810 & 0.334 \\
        \midrule
        Reflow  & 0.080 & 0.070 & 0.000 \\
        RA-Reflow& 0.086 & 0.805 & 0.000 \\
        \midrule
        Reflow+KD & 0.068 & \textbf{0.068} & 0.000 \\
        \modelname{} & \textbf{0.066} & 0.075 & 0.000 \\
        \bottomrule
    \end{tabular}
    }
    \vspace{-0.5cm}
\end{table}

Fig.~\ref{fig:tracking_ablation} provides representative one-step trajectories under the same forward-velocity command, complementing the quantitative results in Tab.~\ref{tab:ablation_tracking}.

\noindent\textbf{Action-flow geometry.}
To further examine the learned transport structure, we visualize the action-flow density of representative Go2 joints in Fig.~\ref{fig:flow_density}.
Across all four legs, the learned trajectories form concentrated transport bands that evolve smoothly over flow time.
This visualization provides a qualitative view of the structured transport induced by \modelname{} and complements the numerical robustness observed in Tab.~\ref{tab:ablation_reward}.

\subsection{Onboard Inference Efficiency}
We evaluate inference latency on the Unitree Go2 onboard CPU using single-thread ONNX Runtime.
As shown in Tab.~\ref{tab:onboard_latency}, reducing the solver budget from 64 steps to a single Euler step decreases mean latency from $4.39$~ms to $0.08$~ms, corresponding to a \textbf{54.9$\times$ speedup}.
The p99 latency simultaneously decreases from $20.06$~ms to $0.10$~ms; the former is approximately the period of a 50-Hz control loop, whereas one-step execution leaves substantial computational headroom.
Together with the preserved one-step return in Tab.~\ref{tab:cross_embodiment}, these results demonstrate the practical deployment benefit of few-step flow control.
\begin{table}[bhtp]
    \centering
    \small
    \caption{\textbf{Onboard inference efficiency on Unitree Go2.}
    Single-thread ONNX Runtime latency on the onboard CPU.}
    \label{tab:onboard_latency}
    \vspace{-0.3cm}
    \resizebox{0.9\linewidth}{!}{
    \begin{tabular}{cccc}
        \toprule
        Euler Steps & Mean (ms) $\downarrow$ & p99 (ms) $\downarrow$ & Speedup $\uparrow$ \\
        \midrule
        64 & 4.39 & 20.06 & 1.0$\times$ \\
        32 & 1.50 & 1.56  & 2.9$\times$ \\
        16 & 0.79 & 1.07  & 5.6$\times$ \\
        8  & 0.41 & 0.44  & 10.7$\times$ \\
        4  & 0.23 & 0.26  & 19.1$\times$ \\
        \textbf{1} & \textbf{0.08} & \textbf{0.10} & \textbf{54.9$\times$} \\
        \bottomrule
    \end{tabular}
    }
\end{table}

\subsection{Real-Robot Deployment}
We deploy \modelname{} on a Unitree Go2 using deterministic zero initialization and fixed one-step Euler execution.
We evaluate three representative locomotion commands: forward locomotion, lateral locomotion, and yaw rotation.
Fig.~\ref{fig:sim2real} shows representative simulation and real-world executions at comparable phases of each motion sequence.

\begin{table}[bhtp]
    \centering
    \small
    \caption{\textbf{Matched-command sim-to-real consistency on Unitree Go2.}
    Joint-position agreement between simulation and hardware using the deployed one-step \modelname{} policy.}
    \label{tab:hardware}
    \vspace{-0.3cm}
    \resizebox{0.8\linewidth}{!}{
    \begin{tabular}{lcc}
        \toprule
        Motion & Joint Pos. Corr. $\uparrow$ & Joint Pos. MAE $\downarrow$ \\
        \midrule
        Forward ($v_x$)   & 0.992 & 0.085 \\
        Lateral ($v_y$) & 0.995 & 0.079 \\
        Yaw Turn ($\omega_z$) & 0.997 & 0.055 \\
        \bottomrule
    \end{tabular}
    }
    \vspace{-0.7cm}
\end{table}

To quantify sim-to-real consistency, we record the command trajectory executed on hardware and replay the same temporally aligned command sequence in simulation.
We then compare the resulting joint-position trajectories.
Tab.~\ref{tab:hardware} reports correlations of $0.992$--$0.997$ and MAE of $0.055$--$0.085$ across the three motions, indicating consistent joint-level responses between simulation and hardware.
This analysis evaluates matched-command joint-position consistency rather than full state-level sim-to-real equivalence.

\section{Limitations}
Although \modelname{} is evaluated across two quadruped and two humanoid embodiments in simulation, our current real-robot validation focuses on Unitree Go2 locomotion.
The hardware evaluation covers a limited set of motion commands, and our matched-command analysis measures joint-position consistency rather than full state-level sim-to-real equivalence.
Broader real-world evaluation on humanoid platforms and more contact-rich tasks would further establish the generality of the approach.
In addition, the deployment efficiency of \modelname{} comes at the cost of additional computation during training.
Online Reflow requires full-budget student rollouts to construct student-induced endpoints, while action distillation relies on a separately trained frozen PPO controller.
Reducing this training overhead and dependence on an external action-space teacher is an important direction for improving scalability.
Finally, our current deployment uses a fixed one-step Euler solver, while adaptive compute regularization is used only during training.
Extending \modelname{} to select the integration budget online according to the current control state may provide a better accuracy--efficiency trade-off for highly dynamic or contact-rich behaviors.

\section{Conclusion}
We presented \modelname{}, a framework for reliable few-step execution of reward-optimized flow policies.
By combining reward-aware online Reflow with complementary multi-budget action supervision, \modelname{} reduces the discretization gap between full-step optimization and low-budget deployment while preserving the original reward-driven objective.
Across two quadruped and two humanoid embodiments, one-step execution closely matches the corresponding 64-step policy, demonstrating robust few-step control across different robot morphologies.
On Unitree Go2, \modelname{} retains \textbf{98.5\%} of the 64-step reward while reducing onboard mean inference latency from $4.39$~ms to $0.08$~ms, yielding a \textbf{54.9$\times$ speedup}.
Real-robot experiments further validate stable one-step locomotion across multiple motion commands.
Overall, these results show that explicitly shaping reward-optimized flow policies for coarse numerical execution provides a practical path toward expressive and efficient robot control.

\addtolength{\textheight}{-12cm}   %

\section*{Acknowledgment}
Generative AI tools were used to assist with language refinement.

\bibliographystyle{IEEEtranS}
\bibliography{IEEEabrv}

\end{document}